\pdfoutput=1

\documentclass[11pt]{article}

\usepackage{ACL2023}
\usepackage{booktabs}
\usepackage{times}
\usepackage{latexsym}

\usepackage[T1]{fontenc}

\usepackage[utf8]{inputenc}

\usepackage{microtype}

\usepackage{inconsolata}

\usepackage{graphicx}

\title{Text Intimacy Analysis using Ensembles of Multilingual Transformers}

  \author{Tanmay Chavan\thanks{~ Equal contribution}~~ \and 
  Ved Patwardhan$\footnotemark[1]$~ \\
  Pune Institute of Computer Technology \\
  \texttt{\{chavantanmay1402, vedpat3\}@gmail.com} }

\begin{document}
\maketitle
\begin{abstract}

Intimacy estimation of a given text has recently gained importance due to the increase in direct interaction of NLP systems with humans. Intimacy is an important aspect of natural language and has a substantial impact on our everyday communication. Thus the level of intimacy can provide us with deeper insights and richer semantics of conversations. In this paper, we present our work on the SemEval shared task 9 on predicting the level of intimacy for the given text. The dataset consists of tweets in ten languages, out of which only six are available in the training dataset. We conduct several experiments and show that an ensemble of multilingual models along with a language-specific monolingual model has the best performance. We also evaluate other data augmentation methods such as translation and present the results. Lastly, we study the results thoroughly and present some noteworthy insights into this problem.

\end{abstract}

\section{Introduction}

Intimacy is a crucial aspect of human nature and communication. The feeling of intimacy implies a certain degree of mutual closeness, openness, and trust between the people involved. It intends to convey information beyond the general semantic meaning, to increase the effectiveness of communication. Capturing this information is a pivotal part of human-like language interaction. Although the intimacy of text is largely determined by its content, several subtle semantic cues can be identified across the text to gauge its intimacy.

The shared task \cite{pei2022semeval} \cite{Pei-EtAl:2023:SemEval} was to predict the intimacy of tweets in ten languages. The primary dataset involved a set of tweets in six languages (English, Spanish, Italian, Portuguese, French, and Chinese) annotated with intimacy scores ranging from one to five. In addition to this data, we used the question intimacy dataset \cite{pei2020quantifying} which contains 2247 English questions from Reddit as well as another 150 questions from Books, Movies, and Twitter with intimacy scores in this dataset ranging from -1 to 1. The model performance has been evaluated on the test set in the given six languages as well as an external test set with four languages absent in the training data (Hindi, Arabic, Dutch, and Korean) based on Pearson's \textit{r} as the evaluation metric. 

Large language models provide a scalable and optimizable framework for performing a large range of text-classification language tasks. Transformers have emerged as the de-facto standard for language modeling due to their ease of parallelization and expressive power as a general-purpose computer. 
Owing to the democratization with pre-trained transformers, we opted for an ensemble to capture distinct sets of features and capabilities. We used bert-based mono-lingual pre-trained models for all six languages as well as multi-lingual pre-trained models to capture inter-language dependencies in the ensemble-based method for this task.

\begin{figure*}[t]
    \centering
    \includegraphics[width=\textwidth]{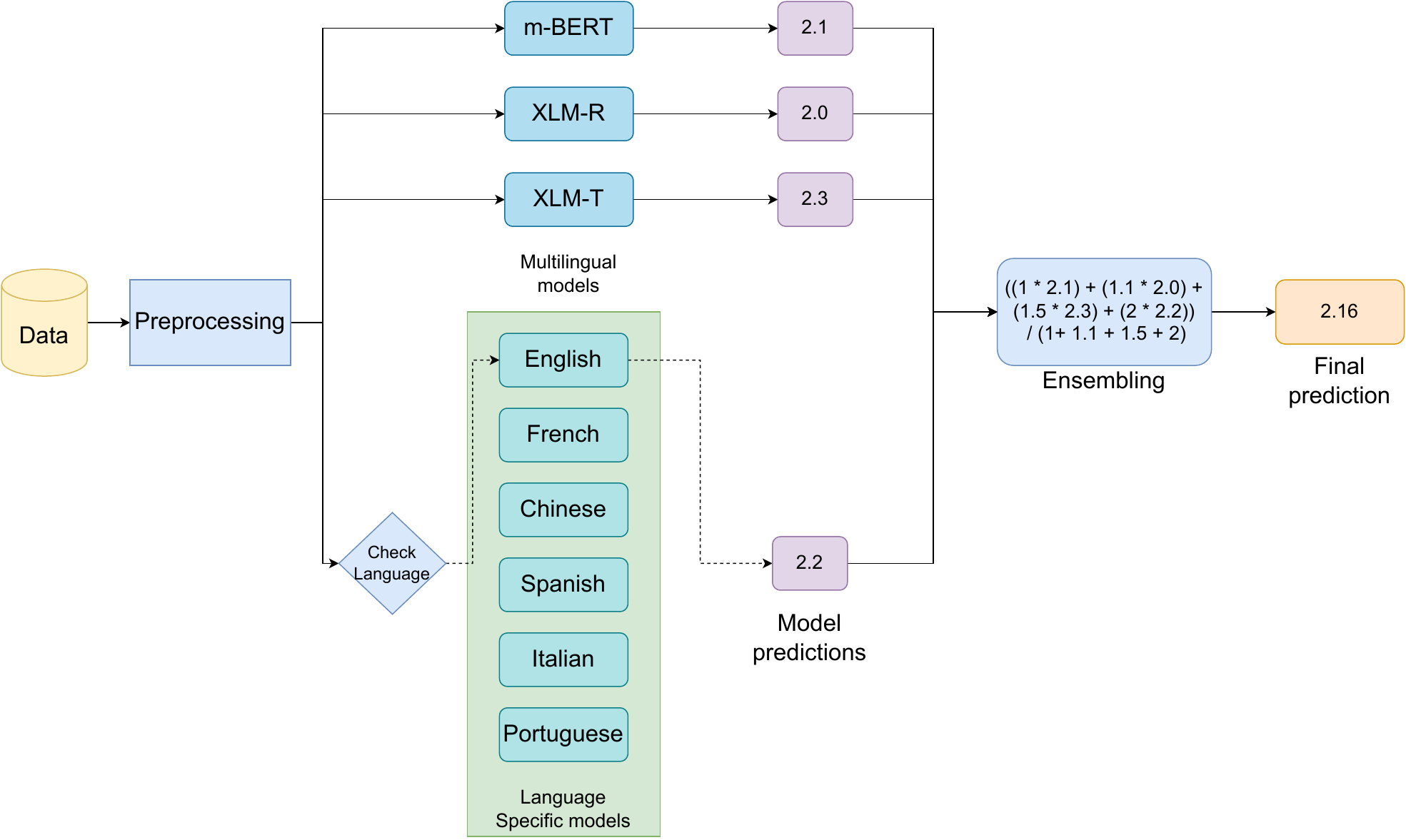}
    \caption{Ensemble system. The final score for a sample is determined as the weighted average of the predicted scores of the individual models.}
    \label{fig:ensemble}
\end{figure*}

\section{Related Work}

The importance of intimacy in our lives is well-studied \cite{Sullivan}. Intimacy, like other interpersonal characteristics, plays an important role in shaping public opinions. Recent efforts have been made to computationally identify such information \cite{choi-etal-2012-hedge}. 

The task of quantifying intimacy in written language was taken up initially by \cite{pei2020quantifying}, where a dataset and model suited to the task were proposed. Data was manually collected from subreddits and annotated using a Best-Worst Scaling (BWS) \cite{flynn2014best} scheme. The model was built on a RoBERTa base and fine-tuned in a self-supervised setting using Masked Language Modelling. MSE and Pearson's \textit{r} correlation were the metrics used for evaluation against human judgment. Their research also suggests interesting conclusions such as the most intimate questions are generally asked between close friends or total strangers.

\section{Data}

The training dataset contains 9491 examples. The testing dataset provided during the submission phase of the task contains 13697 examples. However only 3881 samples are considered in the test set, and the rest of the samples are unannotated and are added to dissuade manual labeling of the data. The unannotated samples have an intimacy score of zero. Thus the actual test dataset only contains 3881 examples. The data consists of tweets obtained from the social networking website Twitter. All of the training samples have intimacy level scores ranging from one to five, where one denotes least intimate while five denotes most intimate. The test dataset also has the same range of intimacy scores.

The task involves predicting the intimacy level of tweets from ten languages. The training dataset consists of tweets from six languages: English, Italian, Portuguese, French, Spanish, and Chinese. The testing dataset, in addition to these six languages, consists of tweets from four more languages: Hindi, Dutch, Arabic, and Korean. The additional languages are absent in the training dataset. They are added to test the capacity of language agnostic performance of models in the prediction of intimacy. All of the tweets are present in scripts native to their languages, although some tweets contain some text from a different script, used for purposes such as acronyms and names. The data is anonymized by replacing username mentions with a generic '@user' handle. The dataset contains hashtags and emojis.

In addition to the given dataset, we used the Reddit questions dataset, containing questions in English from various sources such as tweets, books, and movies in addition to Reddit posts. We used 2247 examples from this dataset. The examples in this dataset originally have intimacy levels ranging from -1 to 1. We mapped these scores to a new range of one to five to be compatible with the main training dataset. We split the training dataset to obtain the training and validation datasets and added the Reddit questions to the training dataset. The training, testing, and validation dataset sizes are summarized in table \ref{tab:split}.

\begin{table}[]
\centering
\begin{tabular}{|c|c|lll}
\cline{1-2}
\textbf{Dataset} & \textbf{Number of examples} &  &  &  \\ \cline{1-2}
Training         & 10029                       &  &  &  \\ \cline{1-2}
Validation       & 1709                        &  &  &  \\ \cline{1-2}
Testing          & 3881                        &  &  &  \\ \cline{1-2}
\end{tabular}
\caption{The train, validation, and test splits of the dataset.}
\label{tab:split}
\end{table}

\section{System Overview}

The presence of multiple languages in the training dataset and additional unseen languages in the testing dataset present unique challenges. We have explored several avenues to solve the problem efficiently. Our approaches utilize pre-trained large language models. We elaborate on our successful as well as unsuccessful experiments in this and the following sections.

\subsection{Ensemble of Multilingual Models}

We experimented with several multilingual models which were trained on all of the languages included in the task. We choose multilingual BERT \cite{devlin-etal-2019-bert}, XLM-RoBERTa \cite{conneau-etal-2020-unsupervised}, and XLM-T \cite{barbieri-etal-2022-xlm} for our system based on their good performance. multilingual BERT includes the top 100 languages of Wikipedia. It is trained on all of the Wikipedia posts on these languages. XLM-RoBERTa also includes 100 languages, however, it uses the larger CommonCrawl corpus for training. XLM-T uses the XLM-RoBERTa model additionally trained on 198 Million scraped tweets with no language restrictions. As XLM-T is trained on tweets in addition to XLM-RoBERTa's prior data, it has superior performance as compared to the other two models for the task.

All of the models are trained on the training dataset. We create an ensemble of these models for predicting intimacy scores on the test dataset. The score for a sample is calculated as the weighted average of the individual predictions of the models. In accordance with their performances, XLM-T has the highest weight, followed by XLM-RoBERTa and multilingual BERT.

\subsection{Ensemble of Multilingual and Language-Specific Models}

Although multilingual models can be used for a wide variety of languages, models trained on a specific language can yield better performance for that language, especially if the language is underrepresented in the training data of the multilingual models. As the tweets individually largely contain only one language, we can use language-specific models as well. These models may be trained purely on one language, or be trained on a specific language on top of the earlier possibly multilingual Thus we use an ensemble of multilingual models along with a language-specific model. We can directly identify the language of an example as it is explicitly mentioned in the training as well as testing datasets. 

We use six different models for each of the languages present in the training dataset. We selected the models based on their performances as well as other characteristics like the similarity of the pre-training data of the model to that of the task. We use twitter-roberta-base-sentiment \cite{barbieri-etal-2020-tweeteval} for English, which is a Twitter-RoBERTa-base model pre-trained on 58 Million tweets and fine-tuned on a sentiment analysis dataset. We use CamemBERT \cite{martin-etal-2020-camembert} for the French Language. It is based on the RoBERTa architecture and trained on a subset containing French text of the OSCAR corpus. We use Chinese BERT (BERT-base-chinese) for Chinese, a BERT model pre-trained on Chinese data. Portuguese BERT (bert-base-portuguese-cased) \cite{souza2020bertimbau} is a BERT model pre-trained on BrWaC, a large corpus in the Portuguese language. twitter-xlm-roberta-emotion-es \cite{vera2021gsi} is an XLM-T model fine-tuned on a Spanish emotion analysis dataset. Italian BERT (bert-base-italian-xxl-uncased) \cite{stefan_schweter_2020_4263142} is a BERT model pre-trained on Wikipedia dumps, the OPUS dataset as well as the Italian subset of the OSCAR corpus.

\subsection{Data Augmentation}

The test dataset contains examples from four languages that aren't present in the training dataset. This implies that the models will predict scores without actually gaining any language-specific information in the training phase. As an experiment, we translate the unseen language samples in the test dataset to English. We then proceed to process it using an ensemble of multilingual models and a language-specific model as mentioned in the above subsection. However, this approach does not yield satisfactory results, which can be seen in section \ref{sec:results}. \\ 

All of our multilingual models were trained with a learning rate of $8e-6$, and the language-specific models were trained with a learning rate of $6e-6$. We used Mean Squared Error as the loss function along with the Adam optimizer. All of the models used in our experiments are openly available on HuggingFace. The HuggingFace model card names are listed in the appendix \ref{sec:hfnames}.

\begin{figure}[]
    \includegraphics[width=\columnwidth]{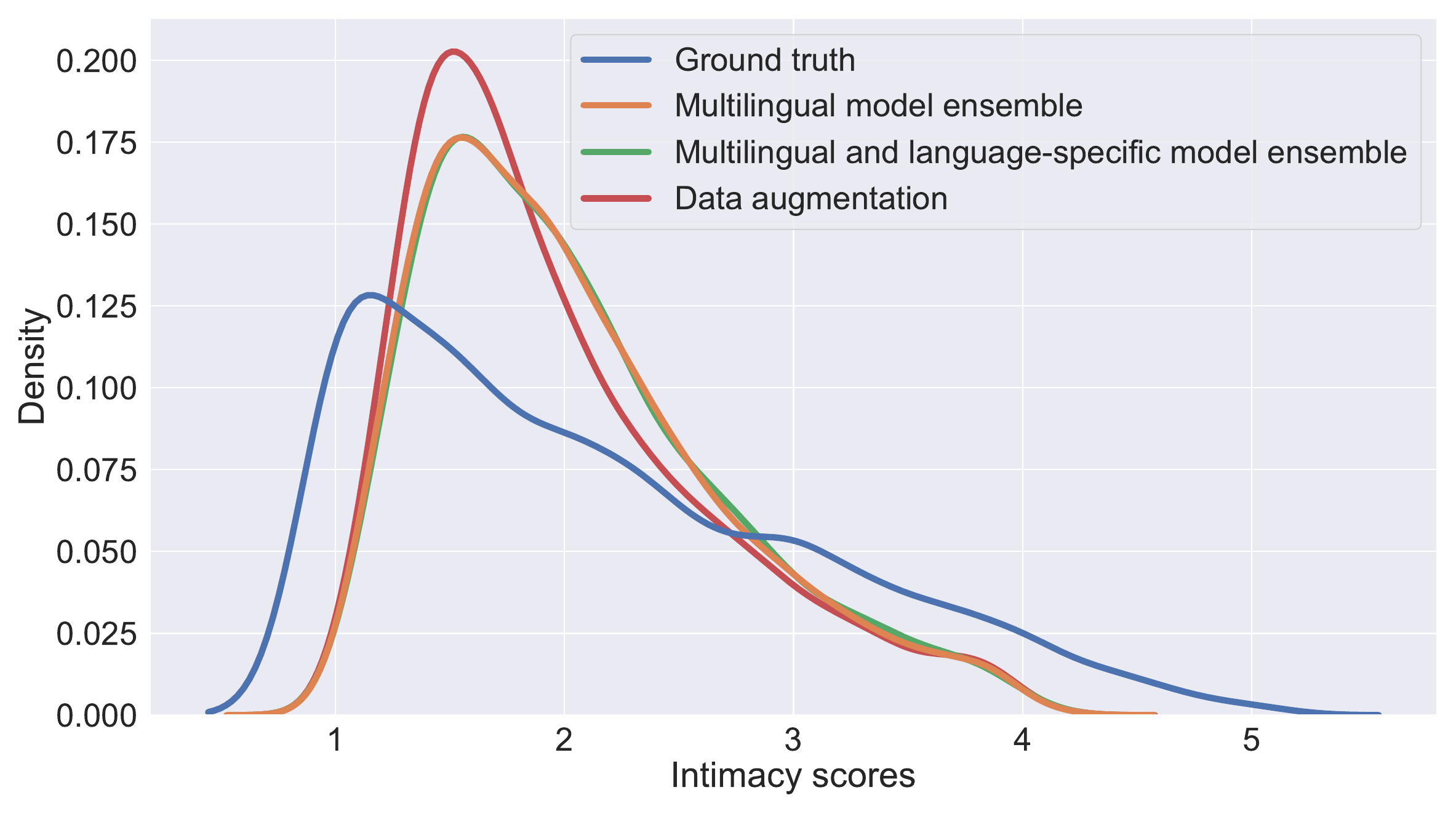}
    \caption{KDE plot of the intimacy scores for the test dataset for all approaches. The multilingual models ensemble and the multilingual models with a lanugage-specific model ensemble have very similar distributions.}
    \label{fig:kdeplot}
\end{figure}

\begin{table*}[]
\centering
\begin{tabular}{|c|c|c|c|l}
\cline{1-4}

\textbf{Language}         & \textbf{\begin{tabular}[c]{@{}c@{}}Multilingual \\ Ensemble\end{tabular}} & \textbf{\begin{tabular}[c]{@{}c@{}}Multilingual and \\ Language-Specific \\ Ensemble\end{tabular}} & \textbf{Data Augmentation} &  \\ \cline{1-4}
English                   & 0.6832  & 0.6911  & 0.6832    &  \\
Spanish                   & 0.7164  & 0.7196  & 0.7164    &  \\
Portuguese                & 0.6636  & 0.6677  & 0.6636    &  \\
Italian                   & 0.7019  & 0.7147  & 0.7019    &  \\
French                    & 0.6894  & 0.7000  & 0.6894    &  \\
Chinese                   & 0.7362  & 0.7431  & 0.7362    &  \\
Hindi                     & 0.1811  & 0.1771  & 0.2053    &  \\
Dutch                     & 0.6250  & 0.6266  & 0.6241    &  \\
Korean                    & 0.3249  & 0.3104  & 0.2242    &  \\
Arabic                    & 0.6149  & 0.6227  & 0.5444    &  \\
\cline{1-4}
\textbf{Overall}          & 0.5697  & \textbf{0.5715}  & 0.5194    &  \\
\cline{1-4}
\textbf{Seen languages}   & 0.7091  & \textbf{0.7154}  & 0.7091    &  \\
\cline{1-4}
\textbf{Unseen languages} & \textbf{0.3972}  & 0.3956  & 0.2884    &  \\ \cline{1-4}

\end{tabular}
\caption{Results of our approaches. The values reported here are Pearson's \textit{r} coefficient.}
\label{tab:results}
\end{table*}

\section{Results}
\label{sec:results}

We present and discuss the results of our experiments in this section. The evaluation metric for the shared task is Pearson's \textit{r}. We have computed the scores using the scorer presented by the organizers. The language-wise, as well as aggregate performances of all our methods, are listed in detail in table \ref{tab:results}.

Our experiments conclude that using an ensemble of various multilingual models along with a language-specific model yields the best overall performance. It has the highest score in the seen languages as well. The ensemble of only multilingual models has the second-best overall score. Note that the multilingual models used in both of these approaches are the same, but with different weights during the ensembling procedure. The data augmentation approach with translation performs worse than the other methods. This approach has inferior performance in all languages, except for Hindi, where the translation seems to provide better results.

It can be observed that our approaches perform similarly well on all of the seen languages, with scores averaging around $0.7$. Among the unseen languages, multilingual models perform considerably poorly in Hindi and Korean texts. However, the models perform substantially better in Dutch and Arabic, despite both of the languages being absent in the training dataset. 

We can see that the multilingual models do not perform equally well on all the languages. This is expected as the pre-training data size varies per language for the models. It is ostensible that the predicted intimacy scores on unseen languages are bound to be less accurate than on seen languages. However, it is interesting to note that certain unseen languages perform considerably better than others. This disparity might suggest two things: the multilingual models perform better on certain languages because they are pre-trained on larger and higher quality corpora of these languages, or the language-agnostic understanding of the models is better for some languages than others. Further research is necessary to disprove the former idea as it is difficult to ascertain the impact of the size of the pre-training data due to the current ensembling process. Also, some models show a disparity in performance on unseen languages despite being pre-trained on similar-sized corpora for these languages.  


The inclusion of a language-specific model in the ensemble provides superior results in every seen language. The increase in performance over the multilingual models also varies per language. This suggests that additional training or fine-tuning of models on specific languages is an effective way of boosting performance for tasks where a sample contains only one language, irrespective of the total number of samples across the dataset.

Translating the dataset yields poor performance than the other methods. This is rather obvious, as translating text often leads to loss of important semantic information. This loss is further exacerbated in tasks where the meaning of the text is important, such as quantifying intimacy in our task. For example, the English word "baby" might be translated into French as \textit{"nourrisson"}, which is more similar to "infant". Although the translation is valid, the original word is far more intimate than its translated counterpart. However, translation yields better performance in the Hindi Language as compared to relying on the multilingual ensemble. This suggests that although translation generally leads to worse results, it might be useful if the model performs very poorly on certain languages.

We submitted the data augmentation-based approach as our final submission for the task. We scored 35th rank as per the overall score on all languages. We finished 18th based on the aggregate score of the seen languages. We also finished 7th in the Chinese language.

\section{Conclusion}

We present our approaches to the SemEval task 9 of multilingual tweet intimacy analysis in this paper. We conducted several experiments involving language models and compared them. We also tried data augmentation methods such as translation of texts from languages absent in the training dataset. We show that an ensemble of multilingual models along with a language-specific model performs well. We show that translation does not yield good results in general. However, other avenues in data augmentation might improve the performance.

\section*{Acknowledgement}

We would like to thank Aditya Kane for his input and for reviewing our work.

\section*{Limitations}

Even though our results are acceptable and might work in the real-life scenario, the current method is deep learning based and does provide analytical guarantees. Moreover, it is noteworthy that each language may pose a different distribution, and a multilingual model might not be sufficient to capture this mixture of distributions. Predicting a single float value from such a mixture of distributions would inevitably be error-prone and noise-sensitive. 



\bibliography{anthology,custom}
\bibliographystyle{acl_natbib}

\appendix

\section{HuggingFace model card names}
\label{sec:hfnames}

\begin{table*}[ht]
\centering
\begin{tabular}{lll}
\cline{1-2}
\textbf{Model name}            & \textbf{HuggingFace model card name}             &  \\ \cline{1-2}
m-BERT                         & bert-base-multilingual-cased                     &  \\
XLM-RoBERTa                    & xlm-roberta-base                                 &  \\
XLM-T                          & cardiffnlp/twitter-xlm-roberta-base              &  \\
CamemBERT                      & camembert-base                                   &  \\
twitter-roberta-base-sentiment & cardiffnlp/twitter-roberta-base-sentiment-latest &  \\
Chinese BERT                   & bert-base-chinese                                &  \\
twitter-xlm-roberta-emotion-es & daveni/twitter-xlm-roberta-emotion-es            &  \\
Portuguese BERT                & neuralmind/bert-base-portuguese-cased            &  \\
Italian BERT                   & dbmdz/bert-base-italian-xxl-uncased              &  \\ \cline{1-2}
\end{tabular}
\caption{HuggingFace model card names for the models used.}
\label{tab:hfmodels}
\end{table*}

\end{document}